\documentclass{article}

\usepackage[preprint, nonatbib]{arxiv}

\usepackage[utf8]{inputenc} 
\usepackage[T1]{fontenc}    
\usepackage{hyperref}       
\usepackage{url}            
\usepackage{booktabs}       
\usepackage{amsfonts}       
\usepackage{nicefrac}       
\usepackage{microtype}      
\usepackage{multirow}
\usepackage[table]{xcolor}
\usepackage{colortbl}
\usepackage{pifont}
\usepackage{siunitx}
\usepackage{pdfpages}
\usepackage{makecell}
\usepackage{graphicx}
\usepackage{longtable}
\usepackage{float}
\usepackage{tcolorbox}
\usepackage{caption}
\usepackage{placeins}
\title{Towards World Models in Biomedical Research}

\author{%
\textbf{Guangyu Wang}\textsuperscript{1,*,$\dag$}\quad
\textbf{Jingkun Yue}\textsuperscript{1,*}\quad
\textbf{Siqi Zhang}\textsuperscript{1,*}\quad
\textbf{Yu Liu}\textsuperscript{2}\quad
\textbf{Xiaoyu Wang}\textsuperscript{3}\quad
\textbf{Mingyuan Meng}\textsuperscript{\textbf{4}}\quad
\textbf{Changwei Ji}\textsuperscript{\textbf{1}}\quad
\textbf{Zongbo Han}\textsuperscript{\textbf{1}}\quad
\textbf{Yulin Wang}\textsuperscript{\textbf{5}}\quad
\textbf{Yang Yue}\textsuperscript{\textbf{5}}\quad 
\textbf{Frank Fu}\textsuperscript{\textbf{6}}\quad
\textbf{Ting Chen}\textsuperscript{\textbf{5}}\quad
\textbf{Song Wu}\textsuperscript{\textbf{3}}\quad
\textbf{Ziwei Liu}\textsuperscript{\textbf{7}}\quad
\textbf{Jiangning Song}\textsuperscript{\textbf{8}}\quad
\textbf{Ming Li}\textsuperscript{\textbf{9}}\quad
\textbf{Gao Huang}\textsuperscript{\textbf{5}}\quad
\textbf{Xiaohong Liu}\textsuperscript{\textbf{3}}\quad
\textbf{Athanasios V. Vasilakos}\textsuperscript{\textbf{10}}\quad
\textbf{Xingcai Zhang}\textsuperscript{\textbf{6,$\dag$}}\quad
\textbf{Ping Zhang}\textsuperscript{\textbf{1}}\quad
\textbf{Yong Li}\textsuperscript{\textbf{5,11,$\dag$}}\\[1em]
\textsuperscript{1}State Key Laboratory of Networking and Switching Technology, Beijing University of Posts and Telecommunications, Beijing, China\\
\textsuperscript{2}Department of Biomedical Engineering, National University of Singapore, Singapore\\
\textsuperscript{3}Institute of Medical Artificial Intelligence, South China Hospital, Medical School, Shenzhen University, Shenzhen, Guangdong, China\\
\textsuperscript{4}Zhongguancun Academy \& Zhongguancun Institute of Artificial Intelligence, Beijing, China\\
\textsuperscript{5}Beijing National Research Center for Information Science and Technology (BNRist), Tsinghua University, 100084, Beijing, China\\
\textsuperscript{6}Department of Chemical and Nano Engineering, University of California, San Diego, La Jolla, CA, USA\\
\textsuperscript{7}Nanyang Technological University, Singapore\\
\textsuperscript{8}Monash Biomedicine Discovery Institute and Department of Biochemistry and Molecular Biology, Monash University, Melbourne, Victoria, Australia\\
\textsuperscript{9}David R. Cheriton School of Computer Science, University of Waterloo, Waterloo, Ontario, Canada\\
\textsuperscript{10}Department of ICT and Center for AI Research, University of Agder (UiA), Jon Lilletuns vei 9, Grimstad, Norway\\
\textsuperscript{11}Department of Electronic Engineering, Tsinghua University, Beijing, China\\[1em]
\textsuperscript{*}These authors contributed equally to this work.\\
\textsuperscript{$\dagger$}Corresponding authors: guangyu.wang24@gmail.com, yuejk@bupt.edu.cn
}

\begin{document}

\maketitle

\begin{abstract}

A central goal of biomedicine is to understand, predict and ultimately control the dynamic mechanisms by which biological systems respond to perturbations, disease progression and therapeutic intervention. Although foundation models and large language models have accelerated biomedical data interpretation, most current systems remain focused on static pattern recognition rather than prospective simulation of biological futures. Here we propose biomedical world models as a paradigm for AI-driven discovery. These models learn latent representations of molecular, cellular, tissue and clinical states, together with intervention-conditioned dynamics that allow future trajectories to be simulated before actions are taken. We discuss how biomedical world models could function as data engines, environment simulators and scientific planning substrates across applications including virtual cells, organoids, virtual patients and surgical simulation. We outline the data infrastructure, evaluation benchmarks, safety constraints and governance frameworks required. Biomedical world models may provide a foundation for simulation-guided, closed-loop and experimentally actionable biomedical discovery. A curated list of relevant papers is available \href{https://github.com/Yuejingkun/Awesome-Biomedical-World-Models}{here}.
\end{abstract}

\section{Introduction}

One central quest of biomedical research is to understand the multiscale mechanisms and dynamics underlying biological systems, enabling their prediction and control under perturbations and interventions. Recent advances in artificial intelligence (AI) have shown potential to accelerate this process, with AI-for-science methods supporting hypothesis generation, molecular design and multimodal data interpretation\cite{gao2024empowering}. More recently, foundation model trained on large-scale multi-modal biomedical data, such as multi-omics, spatial profiling, and imaging, have further enhanced generalization across diverse biomedical domains, enabling a wide range of tasks with limited task-specific labelled data\cite{moor2023foundation,cui2025towards}. However, current biomedical AI primarily focus on correlation modelling and data interpretation\cite{ashenberg2026bridging,rajpurkar2022ai}. Yet biological systems are inherently dynamic, evolving through multiscale processes that respond to genetic perturbations, environmental changes and therapeutic interventions. Understanding such systems therefore requires models that can simulate state evolution rather than merely interpret observations.

In this context, world models provide a promising framework. A defining feature of world models lies in their ability to perform internal “thought experiments” by predicting possible future trajectories before actions are executed. This mirrors long-standing concepts in cognitive science and control theory, where internal models function as a sandbox that support planning, reasoning and decision-making\cite{propoi1963use, richalet1993industrial}. Recent advances in video generation and world models have enabled AI systems to internally simulate physically consistent environments and future scenarios\cite{bruce2024genie,liu2024sora}. In other safety-critical domains such as autonomous driving, world models have shown the potential to play out future traffic situations such as evolving traffic conditions before actions are executed. Building on this foundation, such models enable agents to evaluate candidate interventions and possible future trajectories through internal simulation\cite{du2023learning,bar2025navigation,maes2026leworldmodel,hafner2025training}. Yet, the application of these capabilities to scientific domains, particularly biomedicine, remains largely unexplored.

\begin{figure}[]
\centering
    \includegraphics[width=\linewidth, trim=15pt 280pt 10pt 20pt, clip]{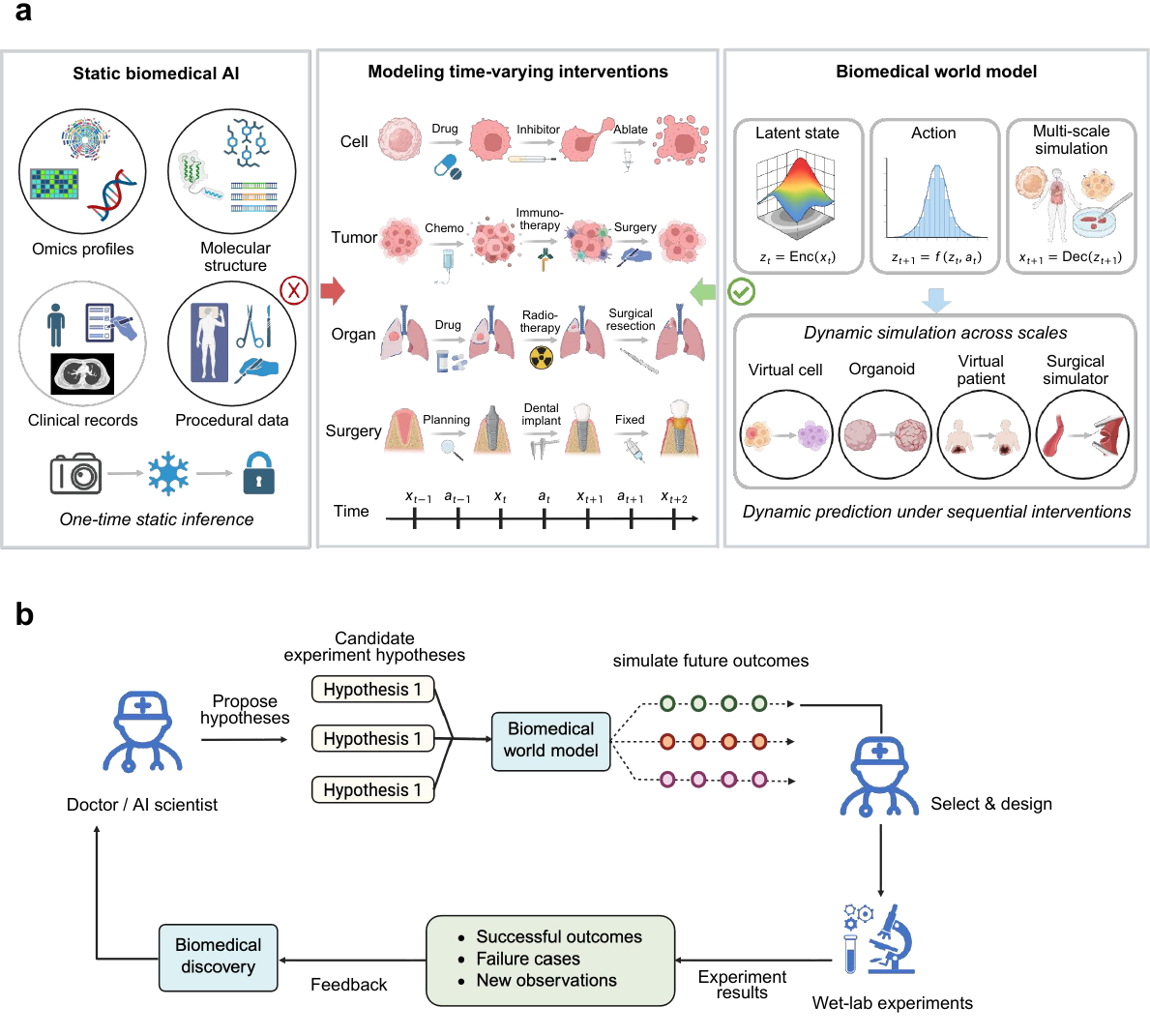}
    \caption{Conceptual framework of biomedical world models.}
    \label{fig:fig1}
\end{figure}

Here, we envision biomedical world models as a promising paradigm for biomedical research. Biomedical world models aim to learn internal representations of biomedical states together with the dynamics governing how these states evolve under perturbations, treatments, and environmental changes (Fig.~\ref{fig:fig1}). Three core capabilities are particularly important. First, biomedical world models can serve as a data engine. Biomedical world models learn multiscale latent representations from heterogeneous and partially observed biomedical data, including omics, imaging, physiological signals and clinical records. By organizing observations into coherent state spaces and generating simulation-enriched trajectories, they may help address data sparsity, improve representation learning and support the development of downstream foundation models. Second, as an environment simulator, biomedical world models seek to capture how biological and clinical states evolve under perturbations. They could support intervention-aware dynamics simulation by enabling the forward simulation of biomedical dynamics under intervention sequences, thereby allowing systematic exploration of competing hypotheses. Third, as a scientific action planner, biomedical world models provide a substrate for closed-loop reasoning and decision-making. Grounded in biologically plausible and mechanistically consistent trajectories, they can support closed-loop scientific reasoning for AI agents. By shifting hypothesis testing and intervention evaluation into in silico environments, biomedical world models could reduce reliance on costly sequential experimentation and enable scalable exploration of the biological hypothesis space.

In this Perspective, we outline the conceptual foundations, opportunities and challenges of biomedical world models. We first define the paradigm and distinguish it from related developments in biomedical foundation models, digital twins, and generative AI. We then discuss representative use cases across biological scales and clinical domains, spanning molecular and cellular dynamics, virtual organoids, precision medicine and surgical simulation. Next, we examine the key ingredients required to build biomedical world models, including multimodal state representations, action-conditioned dynamics and scientific agents that couple simulation with hypothesis generation and experimental planning. Finally, we discuss the major challenges that must be addressed for biomedical world models to become reliable scientific tools, including longitudinal data infrastructure, evaluation and benchmarking, privacy and safety, and large-scale deployment. Ultimately, the promise of biomedical world models is not simply to simulate possible biological futures, but to help scientists and clinicians determine which futures are most informative, plausible and actionable. By bringing simulation into the loop of scientific discovery, biomedical world models may provide a foundation for a new generation of AI systems capable of hypothesis generation, intervention planning and adaptive biomedical discovery.

\section{Overview of World Models}

\subsection{The idea of world models}
The conceptual foundation of world models is closely aligned with long-standing theories in cognitive science and neuroscience that human intelligence relies on internal models to represent latent states of the world\cite{craik1967nature,rao1999predictive}. Early work by Craik\cite{craik1967nature} proposed that intelligent behavior relies on “small-scale models” that allow organisms to anticipate outcomes before acting. Predictive coding and Bayesian brain theories further developed this view, suggesting that perception and cognition arise from internal generative models that integrate sensory inputs with prior beliefs to infer latent states of the world\cite{rao1999predictive,clark2013whatever,friston2010free}. In machine learning, world models were initially explored in reinforcement learning, where transition models are learned to support planning through simulation\cite{sutton1998reinforcement}. More recently, advances in large-scale generative modeling have expanded this idea beyond traditional control settings to encompass dynamics of digital and physical environments, including video generation models and embodied AI systems. Rather than solving a single task, these models seek to construct an internal simulator that can anticipate how a system may evolve under different actions, interventions, or conditions. 

Formally, a world model is an internal, generative or predictive model that learns latent representations of a system and simulates how its states evolve over time under context and, where applicable, actions or interventions. Given historical observations $O_{\le t}$, the model infers a latent state $z_t = f_{\theta}(O_{\le t})$ that summarizes the current state of the system. Conditioned on previous actions or interventions $a_{<t}$, and contextual variables $c$, it then predicts future state transitions through $p(z_{t+1} \mid z_t, a_t, c)$, capturing how the system may evolve under a specific action or intervention. The predicted latent state can be decoded into future observations $o_{t+1}$ through an observation model, $p(o_{t+1} \mid z_{t+1}, c)$, or used to predict task-specific outputs $y_{t+1}$ such as adverse events via an outcome model, $p(y_{t+1} \mid z_{t+1}, c)$. A defining characteristic of world models is their ability to to learn action- or intervention-aware transition dynamics, thereby enabling simulation, long-horizon prediction, counterfactual reasoning, and decision-oriented planning. 

Recent developments suggest that world models are not a single architecture but a spectrum of approaches that differ in how they represent observations, model dynamics and support decision-making. We outline three representative classes.

\paragraph{(1) Sensory-space world models.} Sensory-space generative model world models learn dynamics directly in the sensory space through large-scale generative modelling. Typically implemented using diffusion models or autoregressive transformers on video and multimodal data, they generate temporally consistent and controllable observations conditioned on text, actions or latent action representations, without explicitly modeling the underlying state transitions. Representative examples include large-scale video generation models such as Sora\cite{liu2024sora}, interactive environment models trained from unlabeled video such as Genie\cite{bruce2024genie}, and action-conditioned simulators in domains such as GAIA-1\cite{hu2023gaia}. These models excel at generating realistic and controllable observations, making them attractive as general-purpose simulators. However, consistency in observation space does not necessarily imply consistency with underlying physical mechanisms or causal structure, which remains a key limitation for biomedical research.

\paragraph{(2) Latent-space world models.} A complementary paradigm learns dynamics in compact representation spaces rather than reconstructing observations at every step. JEPA-style models\cite{lecun2022path} exemplify this direction by predicting missing or future representations instead of pixels, aiming to filter out task-irrelevant variability and focus model capacity on abstract structure (V-JEPA2\cite{assran2025v}, DINO-WM\cite{zhou2024dino}). Such models can subsequently be conditioned on actions for planning and control, enabling efficient long-horizon prediction while avoiding the computational cost of pixel-level generation. Prior knowledge can also be incorporated directly into latent dynamics. Approaches such as neural ordinary differential equations, physics-informed neural networks and neural operators, embed governing equations and continuous-time dynamics into the state-transition process\cite{lu2019deeponet}. Such models could offer improved interpretability, extrapolation and scientific plausibility. This capability is particularly relevant for biomedical applications, where biological and physiological mechanisms impose strong constraints on plausible system trajectories\cite{ahmadi2026physics}.

\paragraph{(3) Agent-coupled world models.} World models can also be embedded within planning or policy-learning loops. In this setting, the model serves not only a predictor of future states, but also an internal simulator that allows an agent to evaluate candidate actions through simulated trajectories. Dreamer-style agents represent this paradigm by learning behaviors inside latent world models, typically using recurrent state-space models (RSSMs)\cite{hafner2019learning} that allow future trajectories to be unfolded entirely in latent space. For example, TD-MPC2\cite{hansen2024td} performs trajectory optimization in the latent space of an implicit world model for continuous control, while IRIS\cite{micheli2022transformers} formulates world modeling as sequence modeling with transformers to enable sample-efficient learning from imagined rollouts. For biomedical applications, this paradigm suggests a shift from passive prediction to decision-oriented modelling, in which models help identify which interventions, measurements, or experiments are most informative for treatment planning or scientific discovery. 

\subsection{Defining Properties of World Model in Biomedicine}
Despite substantial progress in modelling natural environments, the role of world models in scientific discovery remains underexplored. Biomedical world models can serve as executable simulators of evolving biological and clinical states. Rather than molecular perturbations or environmental exposures as static variables, they aim to model how such actions reshape latent biological states over time, enabling simulation of alternative futures before experimental or clinical implementation. 

This distinction introduces three defining properties of biomedical world models. First, they must support multiscale latent-state modelling under partial observability. This requires integrating heterogeneous, temporally irregular and often weakly aligned multimodal data into coherent latent state representations that remain biologically meaningful. Second, they should capture intervention-aware dynamics modelling. Beyond representing current states, world models seek to learn how biological systems may evolve under molecular perturbations, treatments, or environmental changes. Such capabilities could support the exploration of competing hypotheses, and candidate interventions through internal simulation of biological dynamics. Third, they should support closed-loop scientific reasoning and adaptive experimentation. As new observations become available, the model can be continuously updated and used to guide subsequent measurements, experiments, or interventions. This shift opens the door to intervention-aware discovery with simulation in the loop: rather than alternating between measurement and post hoc analysis, researchers could iteratively update an internal model of biological or clinical state, use simulation to identify informative experiments or treatment options, and refine the model again as new evidence arrives.

\section{Towards Building World Models in Biomedicine}

We therefore propose biomedical world models as internal, learned models of biological and clinical systems that support latent state inference, intervention-conditioned dynamics modeling and prospective reasoning over future trajectories. To fulfil the potential applications described above, world models for biological and clinical systems should possess several key technical properties. We outline key design principles and technical considerations for building such models.

\subsection{Data for building biomedical world models}

Biomedical world models require a fundamentally different data paradigm from current biomedical foundation models. Existing biomedical foundation models have been enabled by large-scale collections of molecular, imaging and clinical data, allowing them to learn statistical regularities and pattern associations across biological and clinical states through representation learning\cite{cui2024scgpt,xiang2025vision,yue2026medsg,zhang2025generalist}. Recent advances in multimodal measurement technologies, including CITE-seq, 10X Multiome, spatial profiling platforms and the growing integration of clinical records with continuous physiological monitoring, are increasingly generating paired multimodal observations. Such data provide complementary views of a shared underlying biological state and enable the learning of unified state representations across molecular, cellular and clinical scales\cite{stoeckius2017simultaneous,gayoso2021joint,ashuach2023multivi}. Emerging technologies for real-time molecular profiling, continuous physiological monitoring, microfluidic experimentation and high-throughput perturbation screening may therefore play a central role in providing the longitudinal, intervention-aware data required for biomedical world modelling.

Building on these advances, biomedical world models seek to learn how biomedical states evolve across time and intervention. This requires moving beyond static observations towards data that reveal state evolution, intervention-driven transitions and feedback-mediated refinement of biological dynamics. First, longitudinal multimodal observations become essential because they provide temporal continuity and allow models to estimate how latent states evolve over time. Serial imaging, longitudinal clinical records, repeated laboratory tests, continuous physiological monitoring and time-resolved molecular profiling are especially valuable in this regard. Large-scale longitudinal resources, such as UK Biobank imaging cohorts have begun to provide such trajectories of patients\cite{sudlow2015uk,shmatko2025learning}. Yet these data remain fragmented across studies, limited in multimodal coverage, and frequently affected by irregular sampling and observational bias. Building biomedical world models may therefore require new infrastructures for temporal alignment, quality control and harmonization across modalities and institutions.

Second, intervention-rich datasets provide critical information for learning action-conditioned dynamics. Genetic perturbations, drug administration, surgical procedures or treatment adjustments\cite{morshid2019machine,zhang2026lingshu} recorded together with timing and clinical context, make it possible to estimate action-conditioned transitions and support counterfactual simulation of alternative intervention trajectories. However, intervention data remain sparse, heterogeneous and often only partially observed, representing one of the major bottlenecks for biomedical world modelling.

Third, closed-loop experimental feedback provides a route for continual model refinement beyond passive observation. In such setting, model-generated hypotheses or decisions are tested through iterative feedback process, including lab-in-the-loop experiments, adaptive clinical monitoring, or human expert correction, and the resulting evidence is used to update the model. Emerging self-driving laboratory systems and scientific-agent platforms have begun to illustrate parts of this paradigm\cite{rapp2024self,szymanski2023autonomous,boiko2023autonomous,swanson2025virtual}. Over time, such feedback-driven infrastructures may allow biomedical world models to evolve from static offline predictors into adaptive systems that continually refine their internal dynamics through interaction with real biological environments.

\subsection{State representation for biomedical world models}

A core component of biomedical world models is the construction of state representations that transform heterogeneous biomedical observations into a shared latent space suitable for downstream dynamics modelling. A natural starting point is the tokenization paradigm widely used in foundation models and LLMs\cite{thirunavukarasu2023large,achiam2023gpt}, where modality-specific units (e.g., words, image patches, audio frames) are converted into embeddings in a shared vector space. In biomedical settings, such units span multiple biomedical scales and are better understood as modality-dependent representations rather than fixed tokens. For example, protein states may be encoded at the residue level, genomic and single-cell data through gene-level or cell-level representations, and clinical data through event-based encodings of diagnoses, laboratory measurements, or treatment phases. Aligning these representations provides a basis for modeling biomedical systems as evolving latent states rather than isolated observations. 

Given the cross-scale nature of biomedical data, we advocate a hybrid tokenization scheme that integrates continuous and discrete representations within a unified state space. Specifically, continuous tokens are suited for dense, high-dimensional signals, such as medical imaging, physiological waveforms, and quantitative molecular measurements, and can be learned via variational autoencoders\cite{kingma2013auto}, diffusion-based encoders\cite{peebles2023scalable}, or pretrained foundation model encoders\cite{simeoni2025dinov3} that compress observations into compact latent embeddings. Discrete tokens are more appropriate for structured, symbolic, or semantic information, including clinical text, diagnostic codes, biological sequences, and categorical state labels such as disease stages or treatment phases, and can be constructed using vocabulary-based tokenization\cite{radford2019language} or vector-quantization approaches such as VQ-VAE-style codebooks\cite{van2017neural}. Such hybrid representations balance abstraction and fidelity, enabling both structured reasoning and fine-grained state modeling. Beyond representation learning, generative models may also help expand the coverage of biomedical state spaces by synthesizing rare, under-sampled or experimentally inaccessible states. For biomedical world models, such methods are most valuable when linked to temporal and interventional information so that generated observations become testable future states rather than static augmentations.

\subsection{Dynamic modeling and simulation for biomedical world models}
Given a unified representation of biomedical world states, the next key component is to model how these states evolve under various interventions, including treatments, genetic perturbations and environmental changes. The objective is not merely to predict future observations, but to simulate biologically plausible trajectories that support hypothesis evaluation, planning and decision-making. Broadly, two complementary approaches can be considered: Observation-space dynamics modelling\cite{liu2024sora} and latent-space dynamics modelling\cite{zhou2024dino}.  

\paragraph{Observation-space dynamics modeling.} Observation-space approaches directly simulate intervention-conditioned changes in observable biomedical measurements. Depending on the application, these outputs may include future images, tumor morphology, tissue organization, spatial molecular patterns, physiological signals, or perturbation-induced transcriptomic profiles. Rollouts are performed by generating future states conditioned on different actions (e.g., drug dosing, surgical interventions) or additional inputs such as language prompts\cite{polyak2024movie}, visual images\cite{hung2023med}, or other contextual signals. Such models are typically implemented through autoregressive\cite{yan2021videogpt}, neural radiance field\cite{mildenhall2021nerf}, or diffusion-based generative architectures\cite{rombach2022high}. 

Their primary advantage is that simulated outcomes remain directly observable and experimentally verifiable, allowing predicted futures to be compared with real biological or clinical measurements. Such simulations are most useful when the generated observations serve as clinically or biologically interpretable readouts rather than merely realistic appearances. Representative applications include simulating tumour evolution by synthesizing post-treatment tumors under specified therapeutic conditions\cite{yang2025medical}, as well as capturing the distribution of cell states and enabling whole-transcriptome prediction under perturbations\cite{zhang2026lingshu}. More broadly, observation-space world models provide a natural interface between simulation and experimentation by generating biologically meaningful future observations that can be directly validated in laboratory or clinical settings.

\paragraph{Latent-space dynamics modelling.} In contrast, latent-space approaches seek to model intervention-conditioned transitions among compact biomedical states rather than reconstructing full observations at each step. Instead of modelling every observable detail, these approaches aim to capture the underlying dynamical structure governing disease progression, physiological adaptation and biological response to intervention. By abstracting away unpredictable details that generative objectives emphasize, such as acquisition artefacts or modality-specific noise, latent-state models focus on predicting how biologically meaningful states over time in the latent space. This approach enables efficient long-horizon simulation, planning and decision-making without operating on raw observations. Recent examples include the Joint-Embedding Predictive Architecture (JEPA) family, such as V-JEPA series\cite{assran2025v}, DINO-WM\cite{zhou2024dino} and PLDM\cite{sobal2026learning}. In biomedical settings, related approaches have shown promise in radiography\cite{yue2025chexworld} and echocardiography\cite{yue2025echoworld}, where latent representations capture multi-scale anatomical structure and domain variations without pixel-level reconstruction, providing a foundation for modeling temporal or action-conditioned dynamics, such as probe guidance. Prior biological and physiological knowledge can also be incorporated through mechanistic constraints, differential equations or conservation laws, allowing latent transitions to remain consistent with known system behavior. 

More broadly, future biomedical world models may need to integrate hierarchical dynamics across biological scales, ranging from molecular conformations and cellular states to tissues, organs and patients. In such systems, interventions act on latent biological states, while simulated trajectories propagate their consequences across scales and time. Observation-space and latent-state modelling should therefore be viewed as complementary rather than competing approaches. Observation-space models provide observable and experimentally verifiable predictions, whereas latent-state models offer compact representations for long-horizon reasoning and planning. The most powerful biomedical world models will likely combine both perspectives, using latent dynamics to generate and evaluate candidate futures while grounding these simulations in observable biological outcomes.

\subsection{Scientific Discovery Agents with World Models}
Building on state representation and dynamical modelling, biomedical world models provide a foundation for a new generation of scientific discovery agents. Recent LLM-based agents have shown potential to accelerate biomedical discovery\cite{gao2024empowering}, including genetic perturbation design\cite{roohani2025biodiscoveryagent}, drug-combination screening\cite{xu2025multi}, and closed-loop experimental design\cite{rapp2024self}. However, most current systems remain fundamentally reactive. While they exhibit capabilities such as task decomposition, tool use, and planning, they remain limited in explicitly reasoning about how biomedical systems may evolve under intervention\cite{yao2022react}. This limitation becomes particularly pronounced in long-horizon biomedical settings, where interventions may introduce delayed, nonlinear or irreversible consequences that unfold across long time horizons\cite{yu2021reinforcement}. In contrast, world models could provide a complementary paradigm by enabling agents to reason over simulated futures before action execution. by introducing explicit lookahead through internal simulation. This allows candidate actions to be evaluated against their potential future outcomes before commitment. Emerging systems, including WebDreamer, WorldCoder and Computer-Using World Models (CUWM), provide early evidence for this principle by improving decision robustness through simulation and lookahead reasoning in complex environments\cite{gu2024your,tang2024worldcoder,guan2026computer}. In biomedical settings, similar principles could allow agents to evaluate candidate perturbations, treatment strategies or experimental protocols before physical execution. 

In this framework, the world model serves as an internal scientific sandbox. Candidate interventions are propagated through the learned dynamics of biological systems to generate future trajectories, enabling agents to compare alternative hypotheses and prioritize the most informative actions.  Planning therefore shifts from selecting the next plausible experiment to reasoning over multiple possible futures and choosing interventions that maximize expected scientific utility. 

A candidate experiment $a_t$ can be evaluated by a world model conditioned on the current latent scientific state $z_t$, prior knowledge and experimental constraints $c$, yielding a predictive distribution over future state $z_{t+1}$, observations $y_{t+1}$:
\[
p_{\theta}(z_{t+1}, y_{t+1} \mid z_t, a_t, c)
\]

The planner agent can then rank candidate experiments with an acquisition function that integrates predicted utility, expected information gain, feasibility, cost, and safety risk.

More broadly, biomedical world models also establish a closed-loop interface between LLM agents, laboratory automation systems and real-world observations. Agents generate hypotheses and intervention strategies, world models simulate their intervention-conditioned consequences, and experimental feedback are used to continuously update both the underlying state representation and the learned dynamics. This creates a continual cycle of simulation, experimentation and model refinement, analogous to the interaction between human scientists and their internal mental models of biological systems. 

Despite this, realizing this vision remains challenging. Scientific environments are characterized by partial observability, distribution shift, sparse feedback and delayed outcomes. Equally critical is how to assess the causal fidelity of learned world models and their impact on downstream planning reliability, particularly when agents are tasked with guiding experimental design or intervention strategies.

\section{Use Cases of World Model across Biomedical Scales}

Biomedical world models have the potential to shift biomedical AI from static recognition toward dynamic simulation. These models aim to represent how biological and clinical systems evolve over time, respond to interventions. Such capabilities could support a new class of applications, from protein dynamics simulation and virtual cell construction to patient-specific trajectory forecasting, treatment planning, and surgical simulation.

\subsection{World models for molecular and cellular dynamics}
Biological function emerges from dynamic processes spanning multiple molecular and cellular scales. Rather than treating these processes as isolated observations, biomedical world models offer a unified framework for learning how biological systems evolve under genetic, chemical and environmental interventions.

\paragraph{At the molecular level.} Protein science has been transformed by deep learning-based models for predicting protein structure and function\cite{watson2023novo,jumper2021highly,abramson2024accurate}. Yet protein behavior is fundamentally dynamic, shaped by conformational fluctuations, molecular interactions, and environment-dependent transitions across cellular contexts, including post-translational modification cascades that reshape the functional ensemble in vivo\cite{guo2016protein,wodak2019allostery}. Consequently, most existing approaches remain centred on static structures or single-point predictions, capturing only limited snapshots of protein state space. Biomedical world models could shift protein modeling toward learned simulators of molecular behaviour under intervention. Rather than predicting a single structure or affinity score, protein world models would learn how protein conformational distribution evolves under perturbations, such as sequence mutation, chemical modification, ligand binding, changes in pH, or molecular crowding conditions. This would enable simulation-guided exploration of protein state spaces before experimental validation\cite{motlagh2014ensemble}. Such models could enable simulation-guided exploration of protein state spaces, supporting state-dependent small-molecule design, allosteric drug discovery, prediction of adaptive resistance mechanisms and simulation of transient protein interaction networks that are difficult to capture through static docking or conventional pipelines alone\cite{copeland2016drug,boehr2009role}. Realizing such models requires linking sequences to perturbation–response observations from simulation (e.g., free-energy perturbation) or experiment, capturing how conformational ensembles evolve under intervention.

\paragraph{At the cellular level.} A long-standing goal in systems biology is to construct a virtual cell, a computational model that captures cellular states, regulatory interactions, and their evolution under genetic or environmental perturbations. Recent data-driven approaches, including single-cell foundation models, have focused on learning high-dimensional representations of cellular states from large-scale omics data. However, the key bottleneck lies in learning reliable transition rules under unseen, combinatorial, and temporally ordered perturbations. Here, we propose to frame the virtual cell as a world-model-based system. In this formulation, biomedical world models learn latent cellular states from multimodal observations, including transcriptomics, epigenomics, proteomics, spatial profiling, and imaging, and model how these states evolve in response to perturbations\cite{hafner2019learning,zhang2026lingshu,bunne2024build,chuai2026towards,ha2018world}. These capabilities align with emerging efforts such as AlphaCell and Lingshu-Cell, which aim to model perturbation-induced cellular dynamics in a generative and intervention-aware manner\cite{zhang2026lingshu}. Therefore, the virtual cell could support prediction of unseen perturbation responses, simulation of cell-state transitions and closed-loop experimental design, allowing researchers to evaluate candidate perturbations in silico.  

More broadly, molecular and cellular world models provide complementary views of biological dynamics across scales. Protein world models focus on how molecular states evolve under perturbation, whereas virtual cells model how these molecular changes propagate through regulatory networks to shape cellular phenotypes and fate decisions. Together, they could enable in silico exploration of intervention trajectories, allowing researchers to evaluate candidate mutations, drugs and perturbations before committing to costly laboratory experiments.

\subsection{Virtual organoids as tissue-scale world models}
Many biological and disease processes emerge not from individual cells alone, but from collective interactions among cells within spatially organized tissues. Human organoids and organ-on-chip systems provide a promising foundation for this next level of biological modelling\cite{clevers2016modeling}. By recapitulating key structural, functional and heterogeneous properties of native tissues\cite{zhao2022organoids}, these systems bridge the gap between molecular and cellular perturbations and emergent tissue-level phenotypes. Compared with conventional two-dimensional cultures, they provide physiologically relevant platforms for studying disease mechanisms, drug efficacy and toxicity, as well as patient-specific treatment responses\cite{xu2018organoid}. 

Biomedical world models could enable the development of virtual organoids\cite{neagu2026cell}, extending cellular world models towards simulation of multicellular dynamics under intervention. By integrating multimodal observations across molecular, cellular, spatial and phenotypic scales, such systems could learn how tissue states evolve under genetic perturbations, environmental changes and therapeutic interventions. These capabilities could support large-scale in silico drug screening, prediction of therapeutic efficacy and toxicity, simulation of tumour evolution and prioritization of candidate interventions before experimental validation\cite{bai2026artificial}. By coupling world-model-based simulation with organoid experiments in a closed-loop framework, such systems could accelerate translational research, improve therapeutic development and enable more systematic exploration of disease and intervention spaces\cite{yakavets2025machine}. 

Recent work on parabiosis, assembloids and organoids broadens this tissue-scale vision by emphasizing connected multicellular systems that capture communication across cell types, tissues and physiological contexts. Such platforms are well aligned with world-model development because they can generate perturbation-response trajectories that are difficult to obtain from isolated two-dimensional culture models. AI-integrated biomaterials and multidimensional printed scaffolds further suggest that future virtual organoids could jointly model biological response, material microenvironment and therapeutic delivery, enabling closed-loop optimization of tissue constructs and intervention regimens. 

\subsection{Virtual patient models for closed-loop precision medicine}
Precision medicine is fundamentally a sequential decision-making problem, in which clinicians repeatedly decide what to do next as a patient’s condition evolves under disease progression and treatment response. However, previous ML-based medical AI systems, ranging from risk prediction to segmentation and prognosis modeling, remain largely focused on one-off predictions from baseline snapshots. Although foundation models and large language models have substantially improved representation learning and clinical reasoning capabilities\cite{thirunavukarasu2023large}, they are not explicitly designed to capture longitudinal intervention-driven patient dynamics. 

To bridge this gap, world-model show potential to construct virtual patients that simulate disease progression, treatment response and patient evolution over time. Several emerging directions are converging toward this vision. Large-scale disease trajectory models such as Delphi, have demonstrated that patient evolution can be represented as evolving clinical states, enabling prediction of long-term disease progression and future health outcomes\cite{shmatko2025learning}. In parallel, the growing concept of medical digital twins has highlighted the possibility of constructing individualized computational replicas that continuously reflect patient-specific physiology and disease status\cite{laubenbacher2024digital,sadee2025medical}. Together, these developments suggest a broader shift from static prediction toward dynamic modelling of patient evolution.

Building on these advances, virtual patient world models extend trajectory modelling toward intervention-aware simulation. Recent studies have begun to show that patient trajectories can be represented as evolving latent clinical states, allowing models to forecast not only what a patient is, but how a patient may change under alternative clinical decisions\cite{yang2025medical}. In oncology, treatment-conditioned world models can simulate how tumour burden, imaging findings and survival trajectories may evolve under different therapeutic regimens, enabling comparison of alternative treatment plans before intervention\cite{ding2025clarity,wang2023optimized}. Similarly, systems such as the Clinical Environment Simulator (CES) can model clinical care as a sequence of interdependent decisions within a digital hospital environment, where each action influences subsequent patient states and future clinical constraints\cite{luo2026clinical}. More broadly, these systems suggest a shift from predicting a single future outcome to simulating multiple possible futures under intervention. Such capabilities may enable clinicians to ask questions that are difficult to evaluate prospectively, such as how a patient's trajectory might differ under alternative treatment schedules, or different combinations of therapies. 

\subsection{World Models for Surgical Simulation and embodied autonomy}
Surgery represents one of the most demanding settings for biomedical world models because successful intervention requires not only perception of the current scene but also anticipation of how anatomy, tissue and instruments may evolve under future actions. Traditional surgical simulators provide an important foundation for training and robotic skill acquisition, but they often rely on handcrafted environments and simplified physical assumptions, limiting their ability to capture the complexity of real surgical procedures. Recent advances in surgical AI, including vision-language models and foundation models trained on surgical videos, have substantially improved scene understanding tasks such as instrument recognition, phase recognition, critical view of safety assessment, and risk-aware decision support\cite{yang2026large,wang2023foundation}. These systems can effectively answer the question of what is happening now, yet they remain limited in addressing a more clinically relevant challenge: what may happen next if a different action is taken? \cite{maier2017surgical}

This limitation reflects a broader gap between surgical perception and surgical autonomy. Unlike many medical prediction tasks, surgery is fundamentally an embodied interaction process in which decisions continuously alter the underlying environment. Small differences in retraction, dissection, clipping or energy delivery may trigger substantially different anatomical responses, procedural trajectories and complication risk\cite{chen2025far}. As a result, autonomous surgical systems require more than scene understanding; they must learn the causal relationships linking operative state, surgical action and future outcomes. 

To address this challenge, biomedical world models may offer a new paradigm for modelling surgery as an evolving, action-conditioned process with multiple plausible future outcomes\cite{he2025surgworld}. In this formulation, surgical procedures are represented as state–action–future state transitions, where instrument-tissue interactions, biomechanical responses, anatomical priors, and adverse events jointly determine future operative trajectories, The goal is not merely to generate realistic surgical videos, but to learn the causal dynamics that govern how surgical actions reshape anatomy and influence downstream outcomes.

Early work points toward this direction. Surgical embodied intelligence simulators have highlighted the importance of interactive simulation environments for policy learning and autonomous skill acquisition, while emerging surgical vision world models, such as SurgWM, have shown that action-conditioned procedural dynamics can be learned from large-scale surgical videos. More recent SurgWorld-style frameworks further integrate world modelling with vision–language–action learning, enabling simulated surgical trajectories to support planning, decision-making and robotic policy optimization\cite{he2025surgworld,koju2025surgical}. Taken together, these advances suggest a broader shift from perception-centric surgical AI toward simulation-driven, and embodied surgical intelligence. 

Rather than only identifying the current operative state, such systems could enable exploration of alternative procedural futures before action execution.  Such capabilities may allow surgeons and robotic systems to evaluate candidate manoeuvres, anticipate complications and reason about recovery strategies in a virtual environment. This capability is particularly important for rare and high-risk scenarios, including abrupt bleeding, anatomical distortion, landmark loss, and unsafe tissue tension. By generating clinically plausible intervention trajectories in silico, surgical world models could expand operative experience beyond conventional apprenticeship, ccelerate skill acquisition and provide a foundation for embodied surgical autonomy.

\section{Challenges and Limitations}

Challenges and limitations remain toward the development and deployment of biomedical world models. Here we highlight several key challenges that warrant further consideration. 

\subsection{Dataset infrastructure and governance}

Data availability and infrastructure constitute a central bottleneck for biomedical world models. 

Unlike current biomedical foundation models, which primarily benefit from increasing sample scale and modality diversity, biomedical world models additionally require data that preserve temporal continuity, intervention records, and state transitions across biological and clinical trajectories. Rather than learning static associations between observations and labels, world models must learn how biological states evolve under perturbations, treatments and environmental changes. 

However, such data remain fundamentally scarce in biomedicine. Patient trajectories are often sparsely and irregularly sampled, fragmented across institutions and incomplete across modalities, making it difficult to reconstruct coherent biological latent dynamics. Moreover, interventional data which are essential for learning how treatments and perturbations reshape future trajectories, remain particularly limited because of ethical constraints, and the practical difficulty of collecting longitudinal observations at scale. 

These limitations underscore the importance of data infrastructure as an enabling factor. In biomedical research, growing data remain highly limited in biomedical research and are typically scattered across studies, institutions and experimental platforms, each employing distinct collection, annotation and storage protocols, complicating the pipeline from data acquisition to model training. Moreover, privacy and regulatory requirements constrain centralized data aggregation. Therefore, global coordination across consortia would be essential for the development of data collection and versatile algorithms. Progress will require coordinated infrastructures capable of capable of linking multimodal observations, intervention records and temporal trajectories across institutions and experimental platforms. Standardized biomedical ontologies and coding systems, such as ICD, SNOMED CT, and RadLex, may help align heterogeneous biomedical observations into interoperable representations of biomedical states.

Agentic AI systems may further assist in constructing such infrastructures by supporting data curation, harmonization and longitudinal integration across fragmented biomedical records. Rather than only automating preprocessing, future agents could help transform isolated biomedical observations into coherent, reusable representations of evolving biological trajectories. Without such infrastructure, biomedical world models will remain limited by fragmented and weakly connected observations of the very dynamics they are intended to model.  

\subsection{Evaluation and benchmarks}

Rigorous evaluation is critical to the development of biomedical world models, requiring diverse benchmarks built on standardized datasets. Biomedical world models are often performed generative, simulation-based, or intervene on dynamic biological or clinical trajectories, producing trajectories rather than single-point estimates. Current evaluation practices in medical AI remain largely centered on predictive accuracy on static benchmarks, providing limited insight into whether a model has learned reliable biomedical dynamics. As a result, strong performance on existing benchmarks does not necessarily imply that a model produces reliable or meaningful biomedical simulations. Evaluation should therefore move beyond single-point accuracy toward assessing whether a model can support reliable reasoning over evolving biomedical states. 

This mismatch calls for multi-dimensional evaluation frameworks. Beyond predictive validity, it is essential to assess not only whether generated trajectories remain biologically and clinically plausible, but also whether biomedical world models can improve downstream scientific discovery and decision-making efficiency. A comprehensive evaluation framework should therefore examine observational realism, state consistency, temporal coherence, interventional accuracy, counterfactual validity and planning utility—that is, whether the model can generate realistic observations, maintain coherent latent dynamics across modalities and time, accurately predict intervention responses, support causally plausible counterfactual reasoning, and ultimately make biomedical discovery and clinical decision-making faster, more accurate and less costly. 

Metric design is also challenging. Many current biomedical evaluation metrics rely on human annotations or existing databases as ground truth, yet these references may be incomplete, noisy or biased by existing knowledge. This creates a paradox: a world model may identify a previously unrecognized trajectory or propose a novel mechanism yet be penalized because the prediction does not align with existing labels or curated databases. Another challenge arises in long-horizon simulation, where small deviations may accumulate over time and gradually drift toward biologically implausible or unsafe states. Effective evaluation must therefore assess not only local predictive accuracy, but also temporal stability, and robustness under intervention. 

Progress will also require community-level benchmarking efforts. Open benchmark suites, shared tasks, hidden test sets and transparent leaderboards can help standardize evaluation and expose failure modes across models. Such infrastructure would help make the development of biomedical world models more comparable, reproducible and trustworthy, while encouraging models that support scientific discovery rather than merely optimized for narrow static benchmarks.

\subsection{Privacy, safety and fairness}
The development of biomedical world models demands careful consideration of privacy and data security, especially when using sensitive patient or experimental data. Failure to ensure adequate protection may lead not only to privacy breaches and regulatory violations, but also to a loss of patient trust and barriers to clinical translation. Privacy risks are not limited to the release of raw data. Large generative models may memorize rare cases, reconstruct identifiable patient trajectories or expose sensitive information through model outputs. These risks may be amplified when biomedical world models are connected to external tools, hospital information systems or autonomous agents. Prompt attacks, insecure interfaces and poorly controlled model access could create additional routes for information leakage. Privacy-preserving strategies, such as federated learning\cite{li2020review}, secure computation, differential privacy, and encrypted data processing, may help mitigate these concerns, particularly by enabling collaborative model training across institutions without centralizing raw patient data. However, these technical solutions must be combined with transparent data governance, auditable access control and clear documentation of data sources, model limitations, and intended use cases.

Safety risks are also more complex than in static biomedical AI. Biomedical world models are expected not only to predict outcomes, but also to simulate future states, estimate intervention effects, and support planning. As a result, errors can propagate through entire chains of reasoning and action rather than remaining confined to a single prediction. A plausible but incorrect disease trajectory may mislead treatment planning, an overestimated perturbation effect may guide uninformative or unsafe experiments, and an agent-coupled world model may execute flawed plans if uncertainty is poorly calibrated. These concerns are particularly important in closed-loop settings, where model predictions influence the next measurement, experiment or intervention.

Fairness presents an equally important challenge. Biomedical datasets are often unevenly distributed across race, ethnicity, geographic regions, institutions, disease subtypes and measurement technologies. Consequently, world models may learn state-transition dynamics that generalize well for some populations while systematically failing for others. Such risks may be more pronounced than in static prediction models because biases in latent-state estimation can accumulate over long-horizon simulations, counterfactual reasoning and intervention planning.

Addressing these challenges will require not only fairness-aware model development, but also rigorous uncertainty quantification, calibration and prospective evaluation across diverse real-world settings. Before biomedical world models are deployed to guide scientific or clinical decision-making, their simulated trajectories, intervention recommendations and downstream outcomes must be systematically validated against experimental and clinical evidence.

\subsection{Scale and deployment}

The development and deployment of biomedical world models are likely to introduce substantial challenges in scale, cost, and infrastructure. Models that simulate long trajectories across multiple modalities and intervention spaces may require large memory, substantial compute, and prolonged training on complex datasets. These requirements can markedly increase the computational burden associated with both training and inference. Planning-oriented applications may be particularly demanding because candidate interventions often need to be repeatedly simulated under uncertainty before execution. As a result, the practical cost of deploying biomedical world models may extend far beyond that of a single predictive model, especially in settings requiring real-time interaction with laboratory instruments, hospital information systems or autonomous experimental platforms. 

The scale of biomedical world models also creates important deployment challenges. Local deployment in hospitals, laboratories or clinical research centers can help preserve data privacy and support low-latency interaction with biomedical environments, but such deployment requires specialized hardware, engineering expertise, and reliable software infrastructure that may not be broadly accessible. These constraints are especially important for low-resource settings, where the potential value of biomedical world models may be high, but the available computational infrastructure is limited. 

To address these challenges, low-resource and efficient modelling techniques will be important. Knowledge distillation\cite{hinton2015distilling,yue2024ded}, model pruning\cite{cheng2024survey}, quantization, sparse computation, adapter-based tuning and other parameter-efficient methods\cite{hu2022lora} may help reduce model size and inference cost while preserving essential simulation and planning capabilities. Training and deploying large biomedical world models may consume substantial energy and contribute to carbon and water footprints. Therefore, environmentally friendly AI should also become a practical consideration. 

The practical implementation of biomedical world models also involves complex software and hardware engineering. Deploying such models depends on standardized and auditable interfaces for operating within biomedical environments. Standardized operation interfaces should therefore specify state representations, available actions, logging mechanisms, rollback procedures, and human oversight points. Such infrastructure may be essential for ensuring that world-model-driven systems remain reproducible, trustworthy and compatible with real biomedical workflows.

\bibliographystyle{plain}
\bibliography{arxiv}

\end{document}